\documentclass[letterpaper, 10 pt, conference]{ieeeconf}  % Comment this line out if you need a4paper

\IEEEoverridecommandlockouts                              % This command is only needed if 
\UseRawInputEncoding
\usepackage{graphics} % for pdf, bitmapped graphics files
\usepackage{epsfig} % for postscript graphics files
\usepackage{mathptmx} % assumes new font selection scheme installed
\usepackage{amsmath} % assumes amsmath package installed
\usepackage{amssymb}  % assumes amsmath package installed
\usepackage[ruled,linesnumbered]{algorithm2e}
\usepackage{algorithmic}
\usepackage{color}
\usepackage{multirow}
\usepackage{upgreek}
\usepackage[T1]{fontenc}
\makeatletter
\let\NAT@parse\undefined
\makeatother
\usepackage{hyperref}
\usepackage{cleveref}

\newcommand{\bcdot}{\boldsymbol{\cdot}}

\title{\LARGE \bf
	Label-supervised surgical instrument segmentation using temporal equivariance and semantic continuity
}

\author{Qiyuan Wang$^{1,2}$, Yanzhe Liu$^{5}$, Shang Zhao$^{2}$, Rong liu$^{5}$ and S. Kevin Zhou$^{1,2,3,4}$% <-this % stops a space
\thanks{$^{1}$School of Biomedical Engineering, Division of Life Sciences and Medicine, University of Science and Technology of China, Hefei Anhui, 230026, P.R. China. $^{2}$Center for Medical Imaging, Robotics, Analytic Computing \& Learning(MIRACLE), Suzhou Institute for Advanced Research, University of Science and Technology of China, Suzhou, Jiangsu, 215123, P.R. China. $^{3}$Key Laboratory of Precision and Intelligent Chemistry, University of Science and Technology of China, Hefei Anhui, 230026, P.R. China. $^{4}$Key Lab of Intelligent Information Processing of Chinese Academy of Sciences(CAS), Institute of Computing Technology, CAS, Beijing, 100190, P.R. China. $^{5}$Faculty of Hepato-Biliary-Pancreatic Surgery, The First Medical Center, Chinese PLA General Hospital, Beijing, China. }%
\thanks{\textit{Corresponding author: S. Kevin Zhou (skevinzhou@ustc.edu.cn);
                You can apply for annotations through \href{https://forms.gle/AZKQBw4HC5uXgVmg8}{https://forms.gle/AZKQBw4HC5uXgVmg8}.}}%
}

\begin{document}
	\maketitle
	\thispagestyle{empty}
	\pagestyle{empty}
	
	%%%%%%%%%%%%%%%%%%%%%%%%%%%%%%%%%%%%%%%%%%%%%%%%%%%%%%%%%%%%%%%%%%%%%%%%%%%%%%%%
	\begin{abstract}
    In robotic surgery, instrument presence labels are typically recorded alongside video streams, offering a cost-effective alternative to manual annotations for segmentation tasks. Label-supervised surgical instrument segmentation (SIS), a weakly supervised segmentation setting where only instrument presence labels are available, remains underexplored due to its inherently ill-posed nature. Temporal information plays a vital role in capturing sequential dependencies, thereby enhancing representation learning even under incomplete supervision. This paper extends a two-stage label-supervised segmentation framework by leveraging the temporal characteristics of surgical videos from three perspectives. First, a temporal equivariance constraint is introduced to enforce pixel-level consistency across adjacent frames. Second, a class-aware semantic continuity constraint is applied to preserve coherence between global and local regions over time. Third, temporally-enhanced pseudo masks are generated from consecutive frames to suppress irrelevant regions and improve segmentation accuracy. We evaluate our method on two surgical video datasets: the Cholec80 cholecystectomy benchmark and a real-world robotic left lateral segmentectomy (RLLS) dataset. Instance-level instrument annotations, sampled at regular intervals and validated by an experienced clinician, provide a reliable basis for evaluation. Experimental results demonstrate that our method consistently achieves favorable performances over state-of-the-art methods. These findings highlight the effectiveness of incorporating temporal constraints into label-supervised frameworks, offering a promising strategy to reduce annotation costs and advance surgical video analysis.
    \end{abstract}

	%%%%%%%%%%%%%%%%%%%%%%%%%%%%%%%%%%%%%%%%%%%%%%%%%%%%%%%%%%%%%%%%%%%%%%%%%%%%%%%%
	\section{INTRODUCTION}
    Automated visual understanding of laparoscopic and robotic surgery videos is essential to enable autonomous procedures and provide advanced intraoperative assistance. At its core, this capability is dependent on accurate surgical instrument segmentation (SIS). In recent years, SIS has evolved rapidly: early conventional methods~\cite{rieke2016real} have been replaced by deep learning-based approaches~\cite{hasan2019u,islam2019learning,isensee2020or}. In addition, segmentation architectures originally developed for natural images such as DETR~\cite{carion2020end} and Mask2Former~\cite{cheng2022masked} have been adapted to the surgical domain, resulting in further improvements~\cite{gonzalez2020isinet,ayobi2023matis,baby2023forks,dhanakshirur2023learnable,zhao2022trasetr}. However, most current methods rely on fully supervised learning, which requires extensive manual annotations to achieve the robustness and precision demanded by clinical applications. To reduce this annotation burden, recent studies have proposed various incompletely supervised learning strategies.
     \begin{figure}[!t]
		\centering
		\includegraphics[width=3.0in]{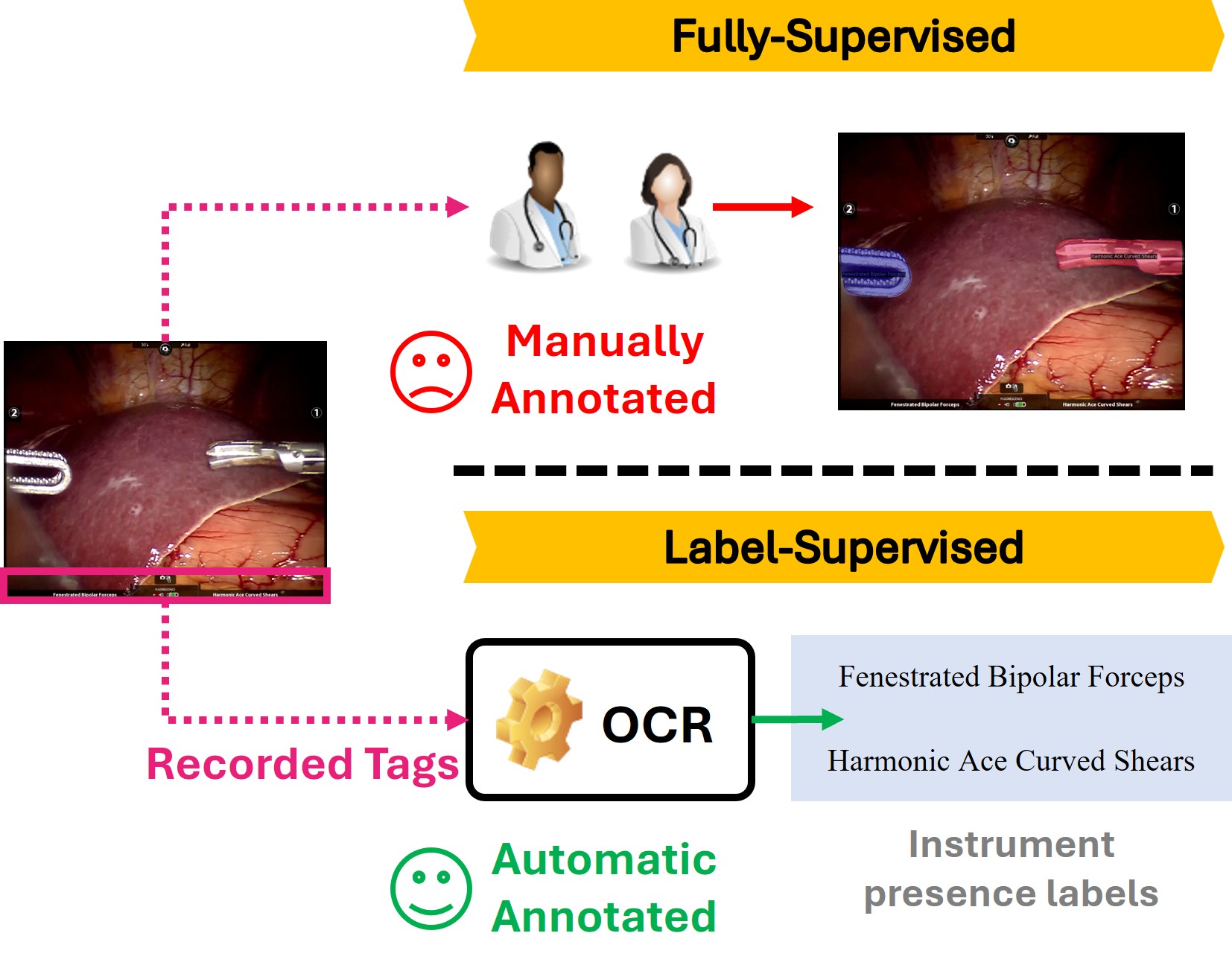}
		\caption{Motivation of label-supervised surgical instrument segmentation which only relies on instrument presence labels. The recorded presence information can be obtained automatically through Optical Character Recognition (OCR) methods at a low cost. This paradigm can alleviate manual pixel-wise annotations for segmentation methods.}
		\label{fig:motivation}
	\end{figure}
    In the unsupervised paradigm, Liu et al.~\cite{liu2020unsupervised} relied on handcrafted cues such as color, objectness, and location to learn implicit segmentation representations. Other methods~\cite{sahu2021simulation,sestini2023fun} used annotated simulated data and applied domain adaptation to real-world surgical scenarios. These approaches can only identify category-agnostic segments and lack semantic information. In the semi-supervised paradigm, Jin et al.~\cite{jin2019incorporating} generated pseudo-masks for unlabeled frames by transferring predictions based on temporal priors. Wei et al.~\cite{wei2023segmatch} extended a widely used semi-supervised classification framework to SIS. Their method applies unsupervised loss to high-confidence pixels in weakly augmented images by contrasting them with adversarially augmented versions. However, these approaches still require a substantial quantity of pixel-level annotations. In the weakly supervised paradigm, Yang et al.~\cite{yang2022weakly} reduced the need for dense segmentation annotations by using scribble-level labels. Matalla et al.~\cite{sanchez2021scalable} proposed a joint detection and segmentation framework based on bounding box annotations and partial presence labels. More recent work~\cite{nwoye2019weakly} has advanced label-supervised localization and detection using only instrument presence information. However, label-supervised surgical instrument segmentation remains an underexplored task. It requires more robust modeling of pixel-level and region-level relationships.

    \addtolength{\topmargin}{0.06in}
    \begin{figure*}[t]
		\centering
		\includegraphics[width=\textwidth]{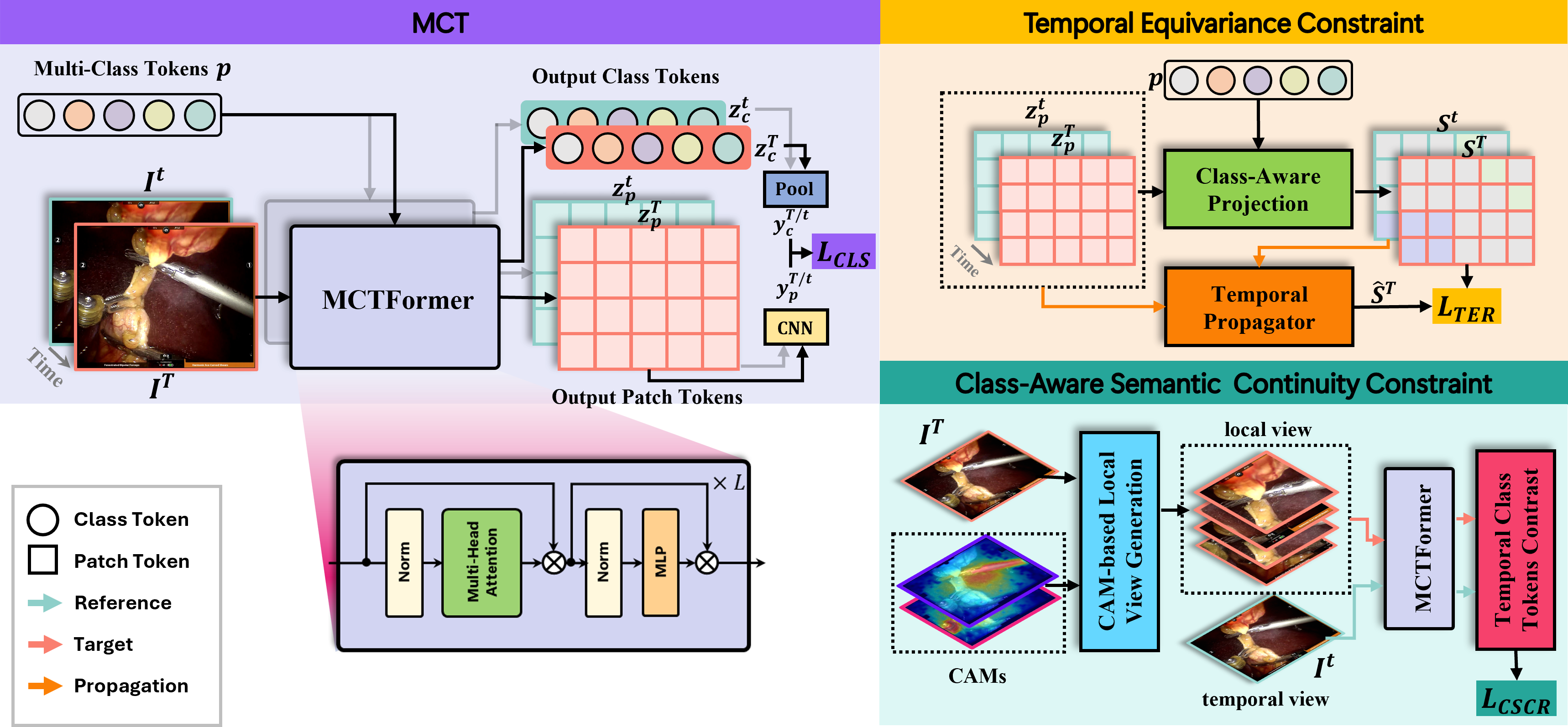}
		\caption{Schematic illustration of an overview pipeline of our method. Two temporality-based constraints are inserted into a two-stage label-supervised architecture.}
		\label{fig:overview}
	\end{figure*}
  
    Robotic surgical videos often record both video streams and instrument presence annotations, offering substantial potential to reduce manual annotation costs, as illustrated in Fig.~\ref{fig:motivation}. However, label-supervised surgical instrument segmentation remains an under-constrained and challenging task. Temporal properties can help alleviate these issues by modeling sequential dependencies and patterns over time under incomplete supervision. To address these challenges, we propose a novel label-supervised surgical instrument segmentation (LS-SIS) framework that relies solely on instrument presence annotations. Our approach incorporates temporality-based constraints to guide the learning process. First, we introduce a temporal equivalence constraint to enhance pixel-level consistency across consecutive features. Second, a class-aware semantic continuity constraint is applied to improve region-level semantic consistency over time for each instrument category. By focusing on uncertain local regions, the model is encouraged to activate non-discriminative parts of the target. Finally, instead of generating frame-wise pseudo-masks, we construct temporal-enhanced masks from a clip of consecutive frames. This strategy helps suppress irrelevant areas while emphasizing foreground regions corresponding to the annotated categories.

    Our main contributions can be summarized as follows: 1) We propose a label-supervised surgical instrument segmentation method, LS-Surg, that relies solely on instrument presence annotations. This approach is well-suited to the surgical domain, where presence annotations are recorded alongside video streams, significantly reducing the cost of manual annotations; 2) We incorporate temporal properties by introducing two temporality-based constraints: temporal equivariance and class-aware temporal semantic continuity constraints, combined with a temporal-enhanced pseudo-mask generation mechanism; 3) Our method demonstrates performance improvements on an open benchmark (Cholec80) and a real robotic left lateral segment liver surgery dataset (RLLS), outperforming prior state-of-the-art methods. Additionally, we contribute instance-wise annotations that were manually labeled and double-checked by an experienced clinician.
    
	\section{METHODS}
    The overall architecture of LS-Surg is depicted in Fig.~\ref{fig:overview}. It is built upon a two-stage label-supervised segmentation framework (LSSS), Multi-Class Token Transformer (MCT)~\cite{xu2022multi}, which was designed for static images. To mitigate highly under-constrained challenges, our method takes temporal attributes of surgical video into account with two temporality-based constraints: temporal equivariance constraint~(\textbf{TER}) and class-aware semantic continuity constraint (\textbf{CSCR}). Furthermore, temporal-enhanced pseudo masks generation (\textbf{TMG}) replaces frame-wise one to suppress irrelevant information.
	
	\subsection{Multi-Class Token Transformer}
    \label{sec:backbone}
    MCT uses multiple class tokens instead of a single one, which is typical in traditional transformers. This design enables class-specific localization by allowing transformer attention to occur between each class token and the patch tokens. To incorporate temporal constraints during training, we extend this architecture to receive temporal pair inputs, as illustrated in the purple section of Fig.~\ref{fig:overview}.

    Given a temporal pair $(I^t, I^T)$, multi-class tokens $p$ aggregate semantic information from corresponding patch tokens through multiple transformer layers. This process yields output class tokens $(z_c^t, z_c^T) \in \mathbb{R}^{D}$ and output patch tokens $(z_p^t, z_p^T) \in \mathbb{R}^{D \times h \times w}$. Here, $D$ denotes the embedding size, while $h$ and $w$ represent height and width, respectively; $t$ and $T$ signify the reference and target time points. These output tokens then follow distinct transformation pathways. The class tokens are average-pooled to generate classification predictions $(y_c^t, y_c^T) \in \mathbb{R}^{C}$, where $C$ is the number of classes. Concurrently, the patch tokens are processed through CNN layers to produce additional class predictions $(y_p^t, y_p^T) \in \mathbb{R}^{C}$. Finally, the multi-label soft margin loss is computed between these class predictions and the ground-truth annotations $(\hat{y}^t, \hat{y}^T)$, as formally expressed in Eq.(\ref{CLS}), where $\sigma$ represents the sigmoid function.
    \begin{equation}
    \begin{split}
        L_{CLS} = \psi(y_c^t,\hat{y}^t) + \psi(y_p^t,\hat{y}^t) + \psi(y_c^T,\hat{y}^T) + \psi(y_p^T,\hat{y}^T) \\
        \psi(y,\hat{y}) = -\frac{1}{C}\sum_{i=1}^{C}\{\hat{y}^i\log \sigma(y^i)+
        (1-\hat{y}^i)\log(1-\sigma(y^i)\}.
    \end{split}
    \label{CLS}
    \end{equation}
    
    We also employ contrastive class-token enhancement, a technique adopted from \cite{xu2023mctformer+}. After the training phase, final class-specific localization maps are generated by combining two key components: class-specific attention maps from the transformer encoder and class activation maps derived from patch tokens. This integration process is detailed in the same aforementioned work. For consistency and clarity, we use the term "CAM" to refer to these composite localization maps throughout this paper.
    
    \subsection{Temporal Equivariance Constraint}
    The \textbf{TER} aims to enhance pixel-wise temporal consistency among patch tokens across video frames, as illustrated in the orange section of Fig.~\ref{fig:overview}. Conceptually, TER works by transforming features from the reference time $\mathrm{t'} = t$ to the current target time $\mathrm{t'} = T$. This ensures that the features from the reference frame align with those of the target frame. This dense transition operation can be effectively implemented using a Temporal Propagator.

    \underline{Class-Aware Projection}: To explicitly establish connections with semantic properties, this operation aims to cluster patch tokens based on multi-class tokens $p$, which can be conceptualized as semantic prototypes. To account for background regions, we \textbf{incorporate a background token} alongside the original multi-class tokens. Utilizing an optimal-transport clustering algorithm~\cite{caron2020unsupervised}, adjacent patch tokens are clustered online based on these semantic prototypes across temporal views. Given a pair of reference output patch tokens $z_{p}^{t}$ and target output patch tokens $z_{p}^{T}$, projection results $(S^t, S^T)$ are derived following this online clustering process.
    
    \underline{Temporal Propagator}: Considering the challenges of appearance variation and fast motion across surgical frames, we adopt a forward warp scheme. This scheme is based on local spatial similarities and avoids the high computational cost in training process, similar to the approach in \cite{salehi2023time}. This propagator can also be alternatively implemented using optical-flow techniques depending on the specific scenario. Given a pair of output patch tokens $(z_{p}^{t}, z_{p}^{T})$, spatial similarities $\mathbf{Q}$ across time are calculated pixel-wise as formulated in Eq.(\ref{eq:sim}), where $\Theta$ represents the cosine similarity operator. Here, $i$ and $j$ denote token index items.
    \begin{equation}
        \mathbf{Q} = [\Theta(z_{p,i}^{t}, z_{p,j}^T)]_{hw\times hw}
        \label{eq:sim}
    \end{equation}
    
    The Temporal Propagator transfers $S^t$ to $\hat{S}^T$ based on the calculated spatial similarities $\mathbf{Q}$, as shown in Eq.(\ref{eq:propagation}), where $\Phi$ represents the propagation function. A local-window thresholding function $\mathcal{N}$ is applied to constrain short-term spatial smoothness, indicating items within a $k$-sized window around item $j$. Subsequently, TER can be formulated as Eq.(\ref{eq:time}) to regulate temporal equivariance. $g$ denotes the argmax function.
   
    \begin{gather}
         \Phi((z_{p}^t,z_{p}^T), S^t) = \hat{S}^T \nonumber\\
         \hat{S}^{T}(j) = \sum_{i} \frac{\exp(\mathcal{N}(\mathbf{Q}_{ij}))}{\sum_{i} \exp(\mathcal{N}(\mathbf{Q}_{ij}))} S^{t}(i), \quad i \in \{1,.., k\}
        \label{eq:propagation}
    \end{gather}
    % \end{equation}
    \begin{equation}
        \label{eq:time}
        L_\mathrm{TER} = - \sum_{i,j}^{hw} g(\hat{S}^T) \log (S^T). 
    \end{equation}
	
	\subsection{Class-aware Semantic Continuity Constraint}
    For surgical frames, there are commonly some \textbf{non-discriminative} object regions that are challenging to differentiate, which can lead to recognition inconsistency across the temporal dimension. Enhancing region-level semantic consistency across time can help the model focus on these more non-salient regions. To this end, we introduce the Class-Aware Semantic Continuity Constraint (CSCR). This constraint facilitates the semantic similarities of region-centric representations between local, uncertain target regions and reference global views for each common category, as depicted in the green portion of Fig.~\ref{fig:overview}.
     \begin{small}
     \begin{gather}
     \label{eq:CTSC}
    L_{CSCR} = 
    \sum_{i=1}^{B_{l}}\sum_{j=1}^{B_{g}}
    		\mathrm{I}_{ij}\log\frac{\exp(x_{i}^{l,T} \bcdot x_{j}^{g,t}/\tau)}{\exp(x_{i}^{l,T} \bcdot x_{j}^{g,t}/\tau)+\sum_{j'=1}^{B_{g}} \mathrm{\overline{I}}_{ij'}\exp(x_{i}^{l,T} \bcdot x_{j'}^{g,t}/\tau))} \\
    B_{l} = B\times C \times L, B_{g} = B\times C \nonumber
    \end{gather}
    \end{small}

    \underline{CAM-based Local View Generation}:
    Specified by the derived CAM, $L$ local crops from the target frame $I^T$ are sampled, with a focus on more uncertain regions. The pseudo multi-label annotation for each crop is then estimated based on the activation area ratio of its corresponding category CAM. If this area ratio exceeds a predefined threshold, the respective category is considered positive within that specific crop.
    
    \underline{Temporal Class Tokens Contrast}:
    As mentioned in subsec.~\ref{sec:backbone}, output class tokens contain region-centric semantic information for all categories. From this, we establish CSCR as the multi-label contrastive formulation. This is achieved by minimizing the difference between global reference and local target class tokens. Specifically, the global and local class tokens, derived from batches of global reference views and $L$ local target crops, are first processed into low-dimensional features $x^{g,t} \in \mathbb{R}^{B \times C \times d}$ and $x^{l,T} \in \mathbb{R}^{B \times L \times C \times d}$ through projection heads ($B$ denotes batch size, and $d$ represents the projection dimension).
     
    The CSCR can be formulated as Eq.(\ref{eq:CTSC}), where $\tau \in \mathbb{R}^{+}$ denotes a temperature parameter. $\mathrm{I} \in \{0, 1\}^{B_{l} \times B_{g}} $ and $\mathrm{\overline{I}} \in \{1, 0\}^{B_{l} \times B_{g}}$ represent intra-class matrix and inter-class indicator matrix, respectively. If two items belong to the same category, the corresponding entry in $\mathrm{I}$ is assigned a value of 1, whereas the analogous entry in $\mathrm{\overline{I}}$ is assigned a value of 0.
    
    The overall training loss can be expressed as Eq.\ref{eq:overall}.
    \begin{equation}
         L_\mathrm{overall} = L_{CLS} + L_{TER} + L_{CSCR}.
         \label{eq:overall}
    \end{equation}
       
    \subsection{Temporal-enhanced Pseudo Masks Generation}
    \begin{figure}[t]
		\centering
		\includegraphics[width=3.4in]{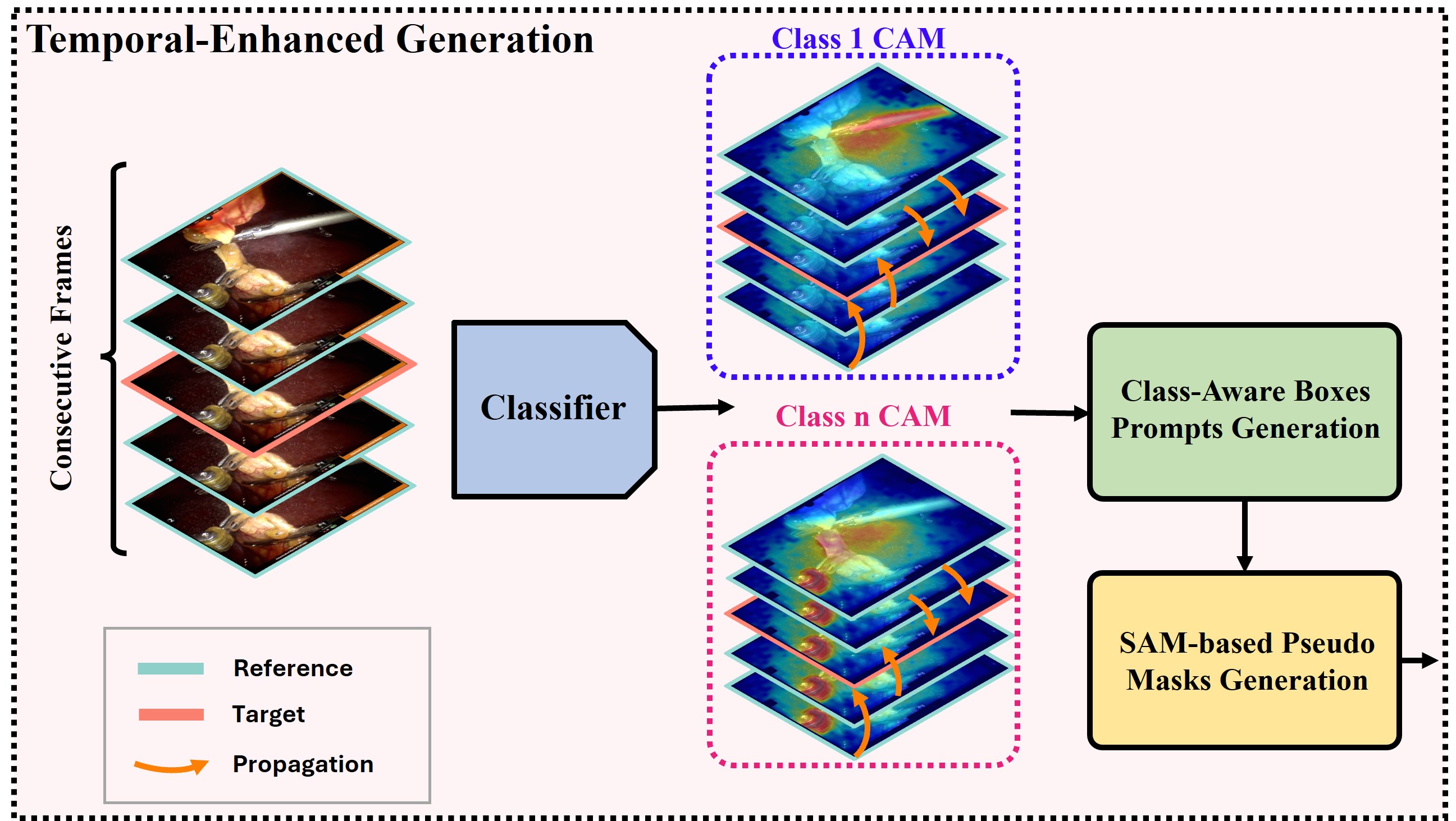}
		\caption{The flowchart of our temporal-enhanced pseudo masks generation.}
		\label{fig:tempral_enhanced}
	\end{figure}
    
    After training, we generate \textbf{temporal-enhanced pseudo masks} rather than frame-wise ones, as depicted in Fig.~\ref{fig:tempral_enhanced}. Given a target frame $I^T$ and its consecutive frames $\{I^{\mathrm{t'}}\}_{\mathrm{t'}=T-\delta t}^{T + \delta t}$, our architecture first derives the output patch tokens $\{z_p^{\mathrm{t'}}\}_{\mathrm{t'}=T-\delta t}^{T + \delta t}$ and subsequently generates their corresponding CAMs $\{M^{\mathrm{t'}}\}_{\mathrm{t'}=T-\delta t}^{T + \delta t}$. CAMs from frames other than $I^T$ are then propagated to time $T$ similar to Eq.(\ref{eq:propagation}). The temporal-enhanced CAMs can be expressed as Eq.(\ref{eq:cam}). This scheme can effectively attenuate background information.
    \begin{equation}
         M^T = \frac{1}{2\delta t +1}(M^T + \sum_{\mathrm{t'} = (T - \delta t)/T}^{T + \delta t} \Phi((z_{p}^{\mathrm{t'}},z_{p}^T), M^{\mathrm{t'}}))
        \label{eq:cam}
    \end{equation}

    Through a thresholding operation, coarse pseudo masks are derived from CAMs, which subsequently generate class-aware box prompts based on connected components. Finally, SAM~\cite{kirillov2023segment} without fine-tuning operation processes class-aware bounding box prompts and produces more precise pseudo masks.

    \subsection{Main Results}
     \begin{figure*}[t]
		\centering
		\includegraphics[width=\textwidth]{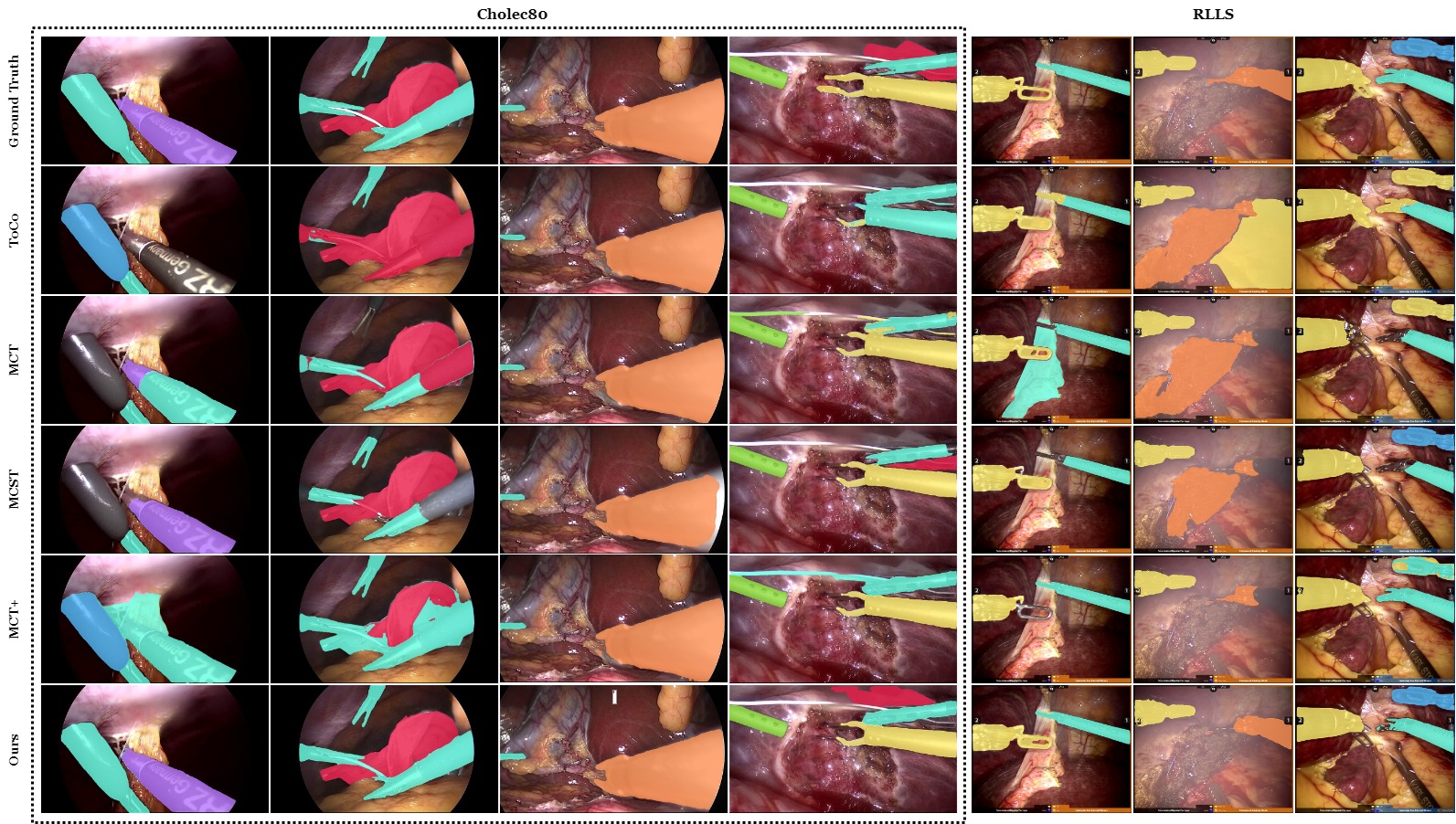}
		\caption{Qualitative comparisons of segmentation networks for both Cholec80 and RLLS. 
        In comparison, our proposed network achieves the best prediction in most situations.}
		\label{fig:weak_rlls}
	\end{figure*}

    \begin{table}[t]
    \caption{Comparison of Ours against different methods on Cholec80,  CholecInstance, and RLLS on semantic segmentation(SS) and instance segmentation(IS) metrics.}
    \label{tab:sota_cholec}
    \begin{center}
    \scalebox{1.}{
    \begin{tabular}{lccccccc}
                \hline
    			\multirow{2}{*}{\textbf{Method}} & \multicolumn{3}{c}{SS} & \multicolumn{3}{c}{IS} \\
                \cline{2-7}
                & $\mathrm{IoU}_{C}$ & $\mathrm{IoU}_{S}$ & $\mathrm{IoU}_{mc}$ & $\mathrm{AP}_{50}$ & $\mathrm{AP}_{75}$ & $\mathrm{AP}$ \\
    			\hline
                \multicolumn{7}{c}{\textbf{Dataset Cholec80}} \\
                \hline
    			WLSTM\cite{nwoye2019weakly}  & 48.17 & 47.55 & 41.72 & 44.24 & 34.15 & 31.82 \\
                MCT\cite{xu2022multi}        & 76.21 & 73.67 & 59.73 & 69.22 & 56.91 & 54.71 \\
                MCT+\cite{xu2023mctformer+}  & \underline{77.17} & \underline{74.87} & 57.77 & 69.23 & 62.20 & 58.11 \\
    			MCST\cite{xu2023learning}    & 71.94 & 69.00 & \underline{66.71} & \underline{79.01} & \underline{67.00} & \underline{61.31} \\
                ToCo\cite{ru2023token}       & 66.79 & 63.99 & 40.10 & 58.31 & 46.21 & 43.87 \\
                \hline
                Ours                         & \textbf{82.08} & \textbf{80.10} & \textbf{73.46} & \textbf{87.73} & \textbf{78.37} & \textbf{73.49} \\
                \hline
    			\multicolumn{7}{c}{\textbf{Dataset CholecInstance}} \\
                \hline
                WLSTM\cite{nwoye2019weakly}  & 50.20 & 49.37 & 37.48 & 33.69 & 23.39 & 22.23\\
                MCT\cite{xu2022multi}        & 69.10 & \underline{66.75} & 52.03 & 58.84 & 47.92 & 43.29\\
                MCT+\cite{xu2023mctformer+}  & 65.94 & 62.77 & 47.09 & 57.67 & 47.87 & 42.90\\
                MCST\cite{xu2023learning}    & \underline{69.11} & 65.13 & \underline{56.12} & \underline{67.85} & \underline{52.03} & \textbf{56.12}\\
                ToCo\cite{ru2023token}       & 65.60 & 62.02 & 35.95 & 51.78 & 40.13 & 36.80\\
                \hline
                Ours                         & \textbf{76.58} & \textbf{72.82} & \textbf{59.35} & \textbf{75.30} & \textbf{62.64}  & \underline{55.98} \\
                \hline
                \multicolumn{7}{c}{\textbf{Dataset RLLS}} \\
                \hline
                WLSTM\cite{nwoye2019weakly}  & 65.53 & 62.68 & 37.95 & 53.27 & 38.66 & 36.78\\
                MCT \cite{xu2022multi}       & 73.42 & 69.76 & 42.36 & 57.48 & 41.58 & 40.49\\
                MCT+\cite{xu2023mctformer+}  & \underline{74.89} & \underline{72.12} & \underline{49.04} & \underline{66.13} & \underline{45.95} & \underline{44.67} \\
    			MCST\cite{xu2023learning}    & 67.71 & 63.27 & 42.55 & 54.11 & 23.68 & 27.19\\
                ToCo\cite{ru2023token}       & 61.59 & 56.99 & 40.35 & 56.86 & 30.00 & 31.08\\
                \hline
                Ours    & \textbf{77.10} & \textbf{73.03} & \textbf{49.89} & \textbf{68.84} & \textbf{53.04} & \textbf{50.23}\\
                \hline
       \end{tabular}%
       }
    \end{center}
    \end{table}
	
	\section{EXPERIMENTS AND RESULTS}
	\subsection{Datasets and Evaluation Metrics}
	\textbf{Datasets}:
   In this paper, we conduct experiments based on two surgical video datasets, which were originally designed for phase recognition tasks, including one cholecystectomy surgery benchmark (\textbf{Cholec80})~\cite{twinanda2016endonet} and one real robotic left lateral segment liver surgery dataset (\textbf{RLLS})~\cite{zhao2022murphy}. Instrument presence labels for RLLS were extracted via Optical Character Recognition (OCR) and noise filtering, while Cholec80 provided those labels. To evaluate LS-SIS methods, we manually annotated instance-wise instrument masks at fixed time steps for both datasets. Specifically, we sampled one frame per minute, excluding \textit{in-vitro} frames, resulting in 1,295 annotated frames for Cholec80 and 2,350 for RLLS. A clinician with three years of experience double-checked all annotations to ensure precision.
   
    \textbf{Metrics}: We evaluate the proposed method from not only semantic segmentation (SS) but also instance segmentation (IS) perspectives. Our research selects the challenge $\mathrm{IoU}_{C}$, $\mathrm{IoU}_{S}$ and $\mathrm{IoU}_{mc}$ metrics defined in~\cite{gonzalez2020isinet} for SS.  The formulas of metrics are represented in Eq.(\ref{eq:metrics}), wherein $P$/$G$ mean segmentation predictions/ground truths, $C$/$\tilde{C_i}$ mean all classes/ground truth classes of the current frame and $K$ represents number of frames. AP metrics in~\cite{cheng2022masked} are applied for IS.
    \begin{small}
    \begin{gather}
    \mathrm{IoU}_{C}\ = \frac{1}{K} \sum_{i=1}^{K} {\left(\frac{1}{\tilde{C_i}}\sum_{c=1}^{\tilde{C_i}} \frac{P_{i,c} \cap G_{i,c}}{P_{ic} \cup G_{i,c}}\right)} \notag \\
    \mathrm{IoU}_{S} = \frac{1}{K} \sum_{i=1}^{K} {\left(\frac{1}{C}\sum_{c=1}^{C} \frac{P_{i,c} \cap G_{i,c}}{P_{ic} \cup G_{i,c}}\right)} \notag \\
    \mathrm{IoU}_{mc} = \frac{1}{C} \sum_{c=1}^{C} {\left(\frac{1}{K}\sum_{i=1}^{K} \frac{P_{i,c} \cap G_{i,c}}{P_{i,c} \cup G_{i,c}}\right)} \label{eq:metrics}
\end{gather}
\end{small}

    \textbf{Implement Details}: Following the two-stage LSSS paradigm, we first train the classification network with instrument presence annotations and generate temporal-enhanced pseudo masks. For Cholec80, the last 40 surgical videos are used for training, while the first 40 videos are reserved for validation. For RLLS, the dataset split aligns with that in~\cite{zhao2022murphy}. Subsequently, a supervised segmentation model is trained with these pseudo masks (excluding validation frames) and then evaluated with ground-truth annotations. The inference time is contingent upon the chosen segmentation model's architecture. In this study, we select Mask2Former~\cite{cheng2022masked} as the segmentor for joint semantic and instance segmentation, achieving a inference speed of approximately $0.01$ seconds per frame (1 A800 GPU).
   
	\textbf{Cholec}: Results of different methods are presented in Table~\ref{tab:sota_cholec}. \textbf{Bold} indicates the optimal results, while \underline{underlined} represents the suboptimal results. In this paper, we select not only some typical LSSS methods like MCT~\cite{xu2022multi}, MCT+~\cite{xu2023mctformer+}, ToCo~\cite{ru2023token}, MCST~\cite{xu2023learning}, but also other similar works in surgical domain like WLSTM~\cite{nwoye2019weakly} for comparison. For SS, LS-Surg outperforms all other methods by a considerable margin. In particular, our method improves over MCT+ by 6\% $\mathrm{IoU}_{C}$, 7\% $\mathrm{IoU}_{S}$ while it improves over MCST by 10\% $\mathrm{IoU}_{mc}$. For IS, LS-Surg is superior to other methods similarly. Ours improves over MCST by 11\% $\mathrm{AP}_{50}$, 16\% $\mathrm{AP}_{75}$, and 19\% $\mathrm{AP}$. 
    
    We also evaluate these trained methods with the validation set of CholecInstance~\cite{alabi2024cholecinstanceseg}, a novel instance segmentation dataset for cholecystectomy surgery, to present the generality of effectiveness. Similar trends can be observed. Following the official dataset split, we also trained an identical segmentation model under supervised learning framework, achieving segmentation metrics of \textit{85.17} $\mathrm{IoU}_{C}$, \textit{81.28} $\mathrm{IoU}_{S}$, \textit{64.58} $\mathrm{IoU}_{mc}$ for SS and \textit{85.78} $\mathrm{AP}_{50}$, \textit{78.76} $\mathrm{AP}_{75}$, \textit{70.28} $\mathrm{AP}$ for IS. The performance gap between ours and supervised one remains within acceptable margins.
    
    \textbf{RLLS}: As presented in Table~\ref{tab:sota_cholec}, for SS, Ours surpasses the sub-optimal method with 2.9\% improvement in $\mathrm{IoU}_{C}$, 1.3\% increase in $\mathrm{IoU}_{S}$ and substantial 1.6\% enhancement in $\mathrm{IoU}_{mc}$. Similarly, in the context of IS, our method improves 4.1\% $\mathrm{AP}_{50}$, 15.4\% $\mathrm{AP}_{75}$ and  12.4\% $\mathrm{AP}$. These metrics validate the importance of temporal properties in surgical video for label-supervised SIS.
    
	\subsection{Ablation Study}
    The ablation study results of key components are presented in Table.~\ref{tab:performance_key_componets}. The first row shows the performance of MCT+. Purely incorporating the temporal equivariance relation via TER, our method significantly improves effectiveness by $5.2\%$ $\mathrm{IoU}_{C}$, $4.4\%$ $\mathrm{IoU}_{S}$, $10.9\%$ $\mathrm{IoU}_{mc}$ for SS and by $20.4\%$ $\mathrm{AP}_{50}$, $19.8\%$ $\mathrm{AP}_{75}$, $21.1\%$ $\mathrm{AP}$ for IS. The CSCR further enhances the baseline across all metrics, with a particularly notable increase in $\mathrm{IoU}_{mc}$. Replacing frame-wise pseudo masks with our TMG continues to boost results, peaking at \textit{82.08} $\mathrm{IoU}_{C}$, \textit{80.10} $\mathrm{IoU}_{S}$ on SS, and \textit{87.73} $\mathrm{AP}_{50}$, \textit{78.37} $\mathrm{AP}_{75}$, \textit{73.49} $\mathrm{AP}$ on IS. The $\mathrm{IoU}_{mc}$ metric, however, shows a slight decrease.
    
    Beyond the final segmentation results, we also evaluate SS on both pseudo masks derived from CAM seeds (PM(C)) and refined pseudo masks from SAM (PM(S)). Given only instrument presence annotations, $\mathrm{IoU}_{C}$ metrics are similar to $\mathrm{IoU}_{S}$. As shown in Table.~\ref{tab:performance_key_componets}, consistent improvement trends are observed across all metrics. Specifically, results show an increase of \textit{11.6} $\mathrm{IoU}_{S}$ and \textit{15.4} $\mathrm{IoU}_{mc}$ for PM(C), and \textit{7.8} $\mathrm{IoU}_{S}$ and \textit{15.5} $\mathrm{IoU}_{mc}$ for PM(S). Compared to PM(S), the method demonstrates a more significant improvement for PM(C).

\begin{table*}[t]
    \caption{Performance of key components on Cholec80. TER means Temporal Equivariance constraint, CSCR means Class-aware Semantic Continuity constraint and TMG means Temporal-enhanced Pseudo Masks Generation. PM(C) represents pseudo masks from CAMs seeds while PM(S) represents refined pseudo masks after SAM. }
  	\begin{center}
 \small
		\begin{tabular}{ccc|rrr|rrr|rr|rr}
			\hline
			\multicolumn{3}{c}{\textbf{Key Componets}} & \multicolumn{3}{|c}{SS} &
			\multicolumn{3}{|c}{IS} & \multicolumn{2}{|c}{PM(C)} & \multicolumn{2}{|c}{PM(S)}\\
   \hline
			TER & CSCR & TMG &  $\mathrm{IoU}_{C}$  &  $\mathrm{IoU}_{S}$  &  $\mathrm{IoU}_{mc}$  &  $\mathrm{AP}_{50}$  &
		  $\mathrm{AP}_{75}$ &  $\mathrm{AP}$  &  $\mathrm{IoU}_{S}$ &  $\mathrm{IoU}_{mc}$  &  $\mathrm{IoU}_{S}$ &  $\mathrm{IoU}_{mc}$ \\ 
			\hline
		            & & & 77.17 & 74.87 & 57.77 & 69.23 & 62.20 & 58.11 & 49.02 & 43.95  & 61.89  & 52.95 \\
			\checkmark & & & 81.15 & 78.15 & 64.11 & 83.38 & 74.55 & 70.37 & 56.29 & 55.94 & 68.13 & 62.39 \\
			\checkmark & \checkmark & & \underline{81.60} & \underline{79.51} & \textbf{73.85} & \underline{84.44} & \underline{76.93} & \underline{71.62} & \underline{60.53} & \underline{57.27} & \underline{69.48} & \underline{65.02} \\
			\checkmark & \checkmark & \checkmark & \textbf{82.08} & \textbf{80.10} & \underline{73.46} & \textbf{87.73} & \textbf{78.37} & \textbf{73.49} & \textbf{60.71} & \textbf{59.44} & \textbf{69.70} & \textbf{68.53} \\
			\hline \\
		\end{tabular}%
      % }\\
	\label{tab:performance_key_componets}
 \end{center}
\end{table*}

 \begin{figure}[!t]
		\centering
		\includegraphics[width=2.9in]{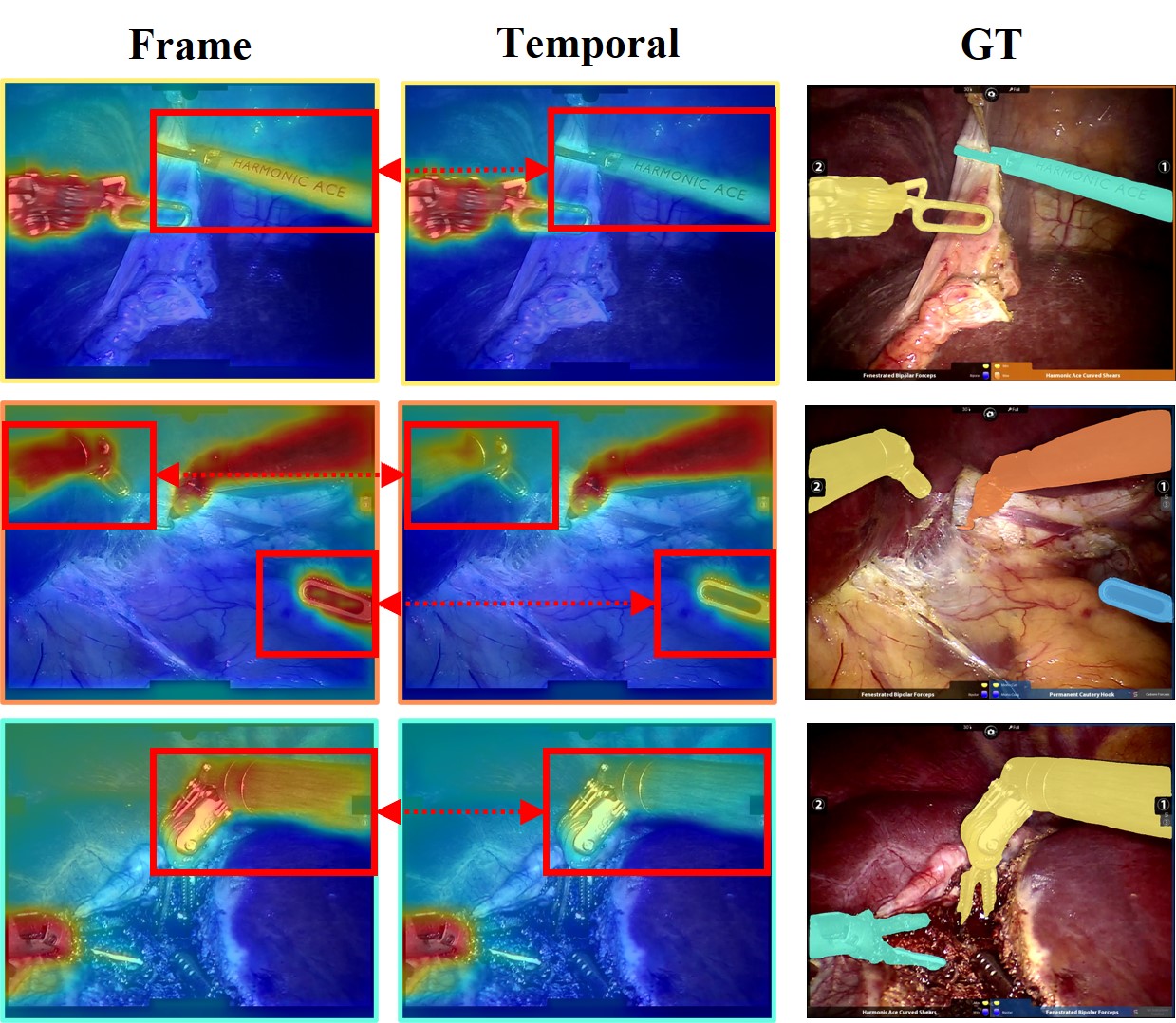}
		\caption{Qualitative comparison for frame-wise and temporal-enhanced CAMs. The red box represents the irrelevant regions that need to be suppressed for the current category.}
		\label{fig:temp_cams}
	\end{figure}
 
    \subsection{Visualization}
    Fig.~\ref{fig:weak_rlls} illustrates the final segmentation results of various methods compared on the Cholec80 and RLLS datasets. Our method consistently demonstrates superior segmentation performance in most cases, exhibiting accurate recognition and comprehensive shape preservation across multiple categories. To further substantiate the effectiveness of the TMG mechanism, we present several examples of temporal-enhanced CAMs alongside their corresponding frame-wise CAMs. As highlighted by the red boxes in Fig.~\ref{fig:temp_cams}, temporal-enhanced CAMs effectively suppress irrelevant regions for the current category. This mechanism enhances focus on category-specific foreground information, thereby improving the accuracy of PM(C) generation.
    
\begin{figure}[!t]
		\centering
		\includegraphics[width=2.8in]{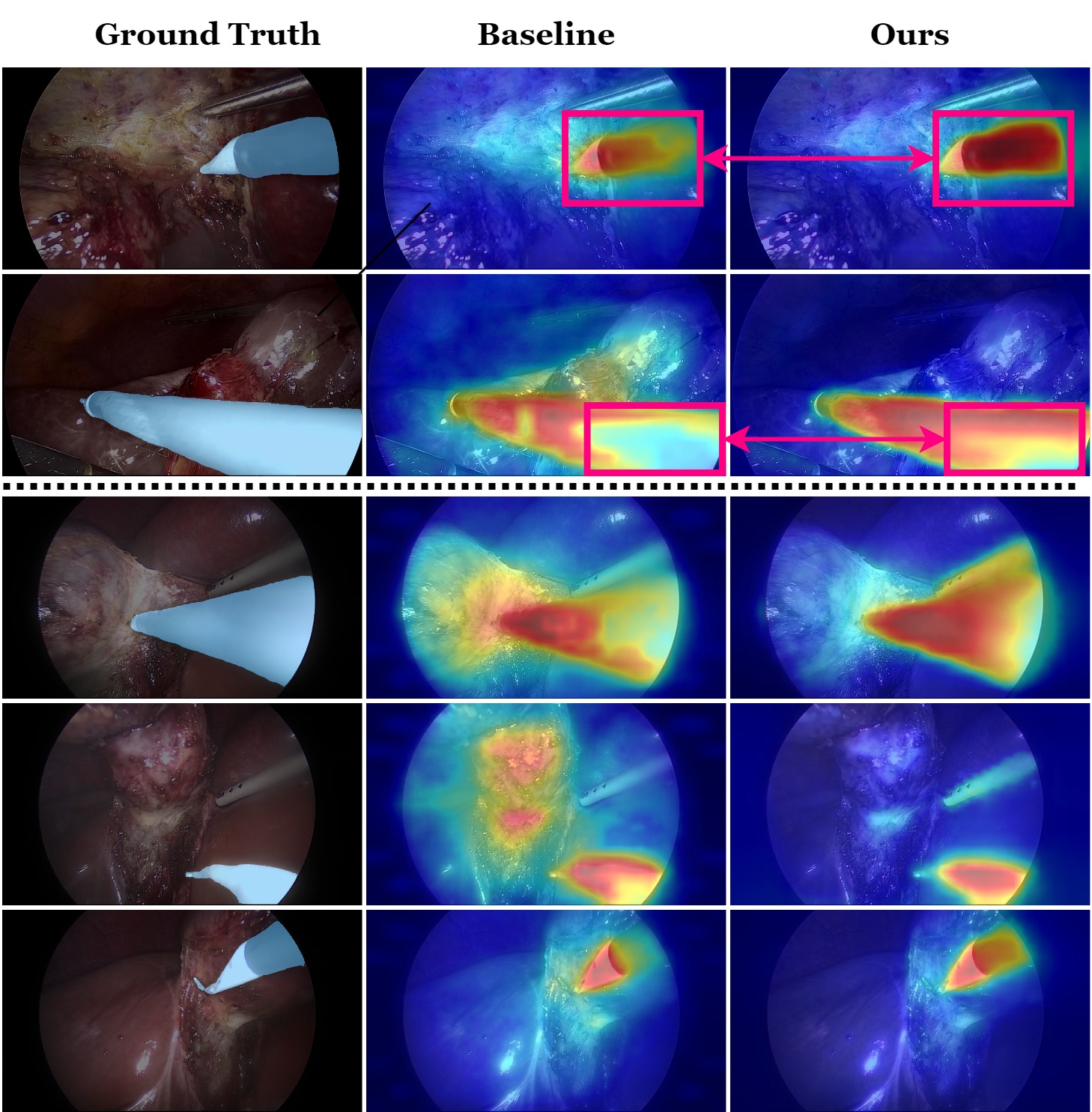}
		\caption{Qualitative comparison of CAMs on low confidence and appearance variation situations. The first two rows represent low confidence regions while the last three rows mark the consecutive three frames with appearance variation.}
		\label{fig:temp_consistency}
	\end{figure}

    In the context of label-supervised SIS, low-confidence regions and appearance variations commonly undermine methodological efficacy. From this, we also visualize the CAMs of both the baseline and our proposed method under these challenging situations to validate robustness. As shown in the first two rows of Fig.~\ref{fig:temp_consistency}, our method can activate more low-confidence object regions than the baseline. For the last three rows, representing three adjacent frames, the CAMs generated by our method demonstrate greater robustness against appearance variations over time.
     
	\section{CONCLUSION}
    
    In this paper, we introduce LS-Surg, a novel approach for label-supervised surgical instrument segmentation (LS-SIS) that relies exclusively on instrument presence annotations. LS-Surg extends a two-stage LSSS framework by incorporating temporal equivariance and class-aware semantic continuity constraints, alongside a temporal-enhanced pseudo-mask generation mechanism. Our extensive experiments on Cholec80 and RLLS datasets validate the effectiveness of LS-Surg. However, the proposed method has several limitations. (1) The lack of tissue presence annotations restricts its applicability to label-supervised tissue segmentation. (2) The two-stage design increases training time. Despite these limitations, our temporal learning strategies can be seamlessly integrated into other frameworks. Future work will focus on adapting them to a single-stage architecture to reduce computational cost and on extending the method to open-vocabulary tissue and organ segmentation using vision-language models. We also plan to explore the use of segmentation outputs in downstream high-level understanding tasks.
   
	%%%%%%%%%%%%%%%%%%%%%%%%%%%%%%%%%%%%%%%%%%%%%%%%%%%%%%%%%%%%%%%%%%%%%%%%%%%%%%%%

	%%%%%%%%%%%%%%%%%%%%%%%%%%%%%%%%%%%%%%%%%%%%%%%%%%%%%%%%%%%%%%%%%%%%%%%%%%%%%%%%

	%%%%%%%%%%%%%%%%%%%%%%%%%%%%%%%%%%%%%%%%%%%%%%%%%%%%%%%%%%%%%%%%%%%%%%%%%%%%%%%%

	\bibliographystyle{IEEEtran}
	\bibliography{root}

\end{document}